\documentclass[journal]{IEEEtran}

\usepackage[utf8]{inputenc}
\usepackage{amsmath}
\usepackage{amssymb}
\usepackage{amsfonts}
\usepackage{graphicx}
\usepackage{subcaption}
\usepackage{booktabs}
\usepackage{multirow}
\usepackage{cite}
\usepackage{xurl}
\usepackage[colorlinks=true,citecolor=blue,linkcolor=blue,urlcolor=blue]{hyperref}
\usepackage{flushend}
\begin{document}

\title{Reliability-Driven Feature Correction for Hyperspectral and LiDAR Classification
}
\author{
Lianming~Zhang, Xiaolu~Wang, Qiya~Song, Jinyang~Liu, Renwei~Dian, Xudong~Kang
\thanks{The authors are with the School of Information Science and Engineering, Hunan Normal University, Changsha 410081, China.}
\thanks{Qiya Song and Renwei Dian are the corresponding authors (e-mail: sqyunb@hnu.edu.cn; drw@hnu.edu.cn).}
}

\maketitle

\begin{abstract}
Hyperspectral image (HSI) and LiDAR data provide complementary spectral and elevation information for land-cover classification.  However, the evidence-based reliability of modality features varies across spatial locations. Less reliable features may adversely affect cross-modal interaction, while adjusting fusion weights alone may not improve the discriminative capacity of these features. To address this issue, we propose ReDF, a reliability-driven framework that quantifies the reliability of each modality's features at each spatial location through the strength of class-related evidence they provide, and uses it to guide feature correction and fusion. Specifically, a Reliability-Guided Feature Correction module introduces a class prototype bank for each modality and converts feature–prototype similarities into class evidence for reliability estimation. The estimated reliabilities are used to construct a reliability-weighted shared representation and to guide bidirectional cross-modal feature correction based on neighborhood agreement, shrinking features with weak class evidence toward this shared reference and refining them with agreement-weighted neighborhood context from the other modality. Building on these outputs, a Reliability-Guided Feature Reweighting module conditions a low-rank modulation on the estimated reliabilities and the cross-modal disagreement, and applies it to the corrected modality-specific descriptors to adaptively rescale each modality's contribution to the fused representation used for final classification. Experiments on three public HSI--LiDAR datasets show higher mean overall accuracy and Kappa than the compared methods. 

\end{abstract}


\section{Introduction}
\label{sec:introduction}

\IEEEPARstart{H}{yperspectral} imagery (HSI) and LiDAR provide complementary observations for land-cover classification, supporting applications in urban planning, agriculture, and environmental monitoring~\cite{ref_survey1,ref_survey2}. HSI captures detailed spectral signatures associated with surface materials, whereas LiDAR provides elevation information that complements spectral observations. Integrating these modalities can improve the discrimination of land-cover categories that are difficult to distinguish using either source alone~\cite{ref_a3clnn,ref_tnnls_psenet}.

Deep learning has advanced HSI--LiDAR fusion through complementary feature extraction and cross-modal interaction. CNN-based methods learn spectral, spatial, and elevation representations through modality-specific branches and coupled feature learning~\cite{ref_cnn1,ref_coupledcnn}. Attention-based and Transformer architectures further enhance multimodal representations by exploiting informative features and spatial context~\cite{ref_tnnls_frit,ref_transformer1,ref_transformer2}. Other approaches, including HFSNet~\cite{ref_hfsnet} and DCU-Net~\cite{ref_dcunet}, organize shared and modality-specific information to retain complementary features. These developments provide effective mechanisms for exchanging and combining information from heterogeneous observations.

However, the evidence-based reliability of modality features varies across spatial locations and land-cover categories. Spectral similarity and mixed pixels can make HSI features ambiguous, while elevation alone may provide limited discrimination between categories with similar heights. Consequently, a modality that is informative in one region may provide weak class evidence in another. Exchanging and fusing features without accounting for this variation may allow weak or ambiguous features to adversely affect the resulting representations. Feature interactions based on cross-modal similarity do not directly establish how much class evidence each modality provides. Explicit reliability estimation is therefore needed to guide the use of complementary information during feature refinement and fusion.

Uncertainty-aware multimodal learning provides a broader context for addressing these limitations. Evidential deep learning estimates uncertainty from class evidence~\cite{ref_evidential1}, while trusted multi-view classification incorporates view-specific evidence into multimodal fusion~\cite{ref_evidential2}. For HSI--LiDAR classification, HFTDF combines feature fusion with trusted decision fusion~\cite{ref_hftdf}, and UCAFNet introduces category-aware weighting~\cite{ref_ucafnet}. Confidence-guided progressive fusion has also been investigated for multimodal classification~\cite{ref_tnnls_tcpfmc}. These studies use uncertainty or confidence estimates to guide the integration of multimodal information. In HSI--LiDAR fusion, adjusting modality weights alone does not refine features with weak class evidence. Reliability-guided cross-modal correction can use complementary observations to refine these features before their contributions are adjusted during fusion.

To this end, we propose ReDF, a reliability-driven framework that uses class-related evidence to guide feature correction and fusion. A Reliability-Guided Feature Correction (RGFC) module maintains a class prototype bank for each modality and converts local feature--prototype similarities into class evidence. The evidence strength provides a reliability estimate for each modality at each spatial location. These estimates weight a shared representation and regulate bidirectional feature correction, which incorporates information from the other modality through neighborhood agreement. This allows complementary observations to refine modality features while accounting for their local reliability.

Building on the corrected features, a Reliability-Guided Feature Reweighting (RGFR) module adaptively adjusts modality contributions to the fused representation. It conditions low-rank modulation on descriptor content, estimated reliabilities, and cross-modal disagreement, and reweights the corrected modality-specific descriptors while preserving the shared descriptor. Thus, the reliability estimates used for feature correction also inform the subsequent fusion of the corrected representations for classification. 
Experiments on three public HSI–LiDAR datasets show that ReDF achieves higher mean overall accuracy and Kappa than the compared methods under the adopted evaluation protocol.

The main contributions are summarized as follows:
\begin{enumerate}
    \item We propose ReDF, a reliability-driven framework for hyperspectral and LiDAR classification that uses prototype-derived evidence to guide feature correction and subsequent fusion.

    \item We develop RGFC to refine modality features using prototype-derived reliability and cross-modal neighborhood agreement, followed by RGFR, which adaptively reweights their contributions to fusion through reliability-conditioned low-rank modulation.

    \item Experiments on three public HSI--LiDAR datasets demonstrate the effectiveness of ReDF, which achieves higher mean overall accuracy and Kappa than the compared methods.
\end{enumerate}

\section{Related Work}
\label{sec:related_work}

\subsection{HSI--LiDAR Fusion for Land-Cover Classification}
\label{subsec:related_fusion}

HSI--LiDAR fusion combines spectral, spatial, and elevation information for land-cover classification~\cite{ref_survey1,ref_survey2}. Early methods integrate extinction profiles with total variation component analysis~\cite{ref_traditional2} or CNN features~\cite{ref_traditional1}. CNN-based methods learn complementary representations directly from paired observations. Xu et al.~\cite{ref_cnn1} combine spectral--spatial HSI features with LiDAR features, while Hang et al.~\cite{ref_coupledcnn} introduce parameter sharing and fusion at both feature and decision levels. Fu et al.~\cite{ref_tnnls_psenet} enhance spatial and spectral semantics for HSI--LiDAR classification. For multisource remote sensing data, Wei et al.~\cite{ref_tnnls_mmqrn} use quaternion representations to integrate multilevel features.

Attention mechanisms and hierarchical architectures further enhance cross-modal interaction. FusAtNet~\cite{ref_attention1} and A$^3$CLNN~\cite{ref_a3clnn} introduce attention mechanisms for multimodal feature learning. Multistage information-complementary fusion~\cite{ref_flexmcfnet} integrates complementary information at successive processing stages. Hierarchical and multimodal fusion Transformers model dependencies across spatial positions and representation levels~\cite{ref_transformer2,ref_transformer1,ref_crosshl}. Other methods combine hierarchical CNN processing with a Transformer~\cite{ref_hct} or adapt fusion across scales~\cite{ref_dshfnet}. Zhao et al.~\cite{ref_tnnls_frit} introduce a fractional Fourier image Transformer to capture local and global context in multimodal remote sensing data.

Feature separation and cross-modal refinement are also used to preserve and exploit complementary information. HFSNet~\cite{ref_hfsnet} employs height-guided fusion and separation, whereas DCU-Net~\cite{ref_dcunet} models common and unique features. Lin et al.~\cite{ref_dcmnet} introduce dynamic routing and bilinear attention to adapt cross-modal feature interactions for HSI--LiDAR classification. In road extraction, Liu et al.~\cite{ref_tnnls_cmmpnet} introduce cross-modal message propagation with dual enhancement to refine modality representations. However, both shared and modality-specific features can provide weak class evidence at individual locations. Separating these components does not directly establish their reliability for classification.

\subsection{Uncertainty- and Quality-Aware Multimodal Fusion}
\label{subsec:related_quality}

Uncertainty and confidence estimation provide a basis for assessing modality information in multimodal fusion. Sensoy et al.~\cite{ref_evidential1} model class probabilities with a Dirichlet distribution to quantify uncertainty associated with insufficient evidence. Han et al.~\cite{ref_evidential2} combine view-specific evidence through Dempster--Shafer theory and subsequently develop dynamic evidential fusion based on sample-level reliability~\cite{ref_tmc_tpami}. Related studies investigate opinion aggregation under subjective logic~\cite{ref_liu_aaai} and conflicts between views~\cite{ref_rcml}.

Modality quality estimates are also used to guide feature interaction and fusion. Zhang et al.~\cite{ref_qmf} analyze dynamic fusion and learn modality weights based on information quality. Cao et al.~\cite{ref_cao_pdf} predict collaborative belief from modality-specific and joint confidence. In multi-target domain adaptation, Huang et al.~\cite{ref_tnnls_emda} derive weights from domain-distribution discrepancies to discount evidence. Li et al.~\cite{ref_tnnls_tcpfmc} combine modality-energy-based confidence evaluation with cyclic progressive fusion, while Liang et al.~\cite{ref_tef} couple fusion architecture search with view enhancement.

Class prototypes provide class-level references in learned
feature spaces, as demonstrated by prototypical networks
for few-shot classification~\cite{ref_protonet}. Liu et al.~\cite{ref_rpcl_fsl} investigate refined prototypical contrastive learning for few-shot HSI classification. For multimodal medical image segmentation, Huang et al.~\cite{ref_huang_evidential,ref_huang_infofusion} map features to voxel-wise evidence using prototypes and apply contextual discounting before evidence combination. Their reliability coefficients are learned for each modality--class pair and act on the evidence supplied to fusion.

In remote sensing, Zhao et al.~\cite{ref_reliability_rs1} use evidential uncertainty to weight aerial and ground views. For HSI--LiDAR classification, HFTDF~\cite{ref_hftdf} combines shallow and deep feature fusion with trusted decision fusion. UCAFNet~\cite{ref_ucafnet} weights the spectral, spatial, and elevation branches according to their per-class classification accuracy and also purifies and enhances shared spatial features. Wang et al.~\cite{ref_udfnet} further investigate uncertainty-aware dynamic fusion for multimodal
land-cover classification, assigning modality weights
according to uncertainty measured by energy scores.

These studies incorporate uncertainty and information quality into different stages of multimodal fusion. However, evidence discounting and category-dependent branch weighting do not by themselves link local class support to cross-modal feature correction and the subsequent fusion of corrected representations. ReDF addresses this issue by deriving local modality reliability from the strength of class-prototype evidence and using it to guide both feature correction and the weighting of corrected descriptors.

\section{Methodology}
\label{sec:methodology}

\subsection{Overall Framework}
\label{subsec:overall_framework}

\begin{figure*}[!t]
    \centering
    \includegraphics[width=\textwidth,pagebox=cropbox,clip]{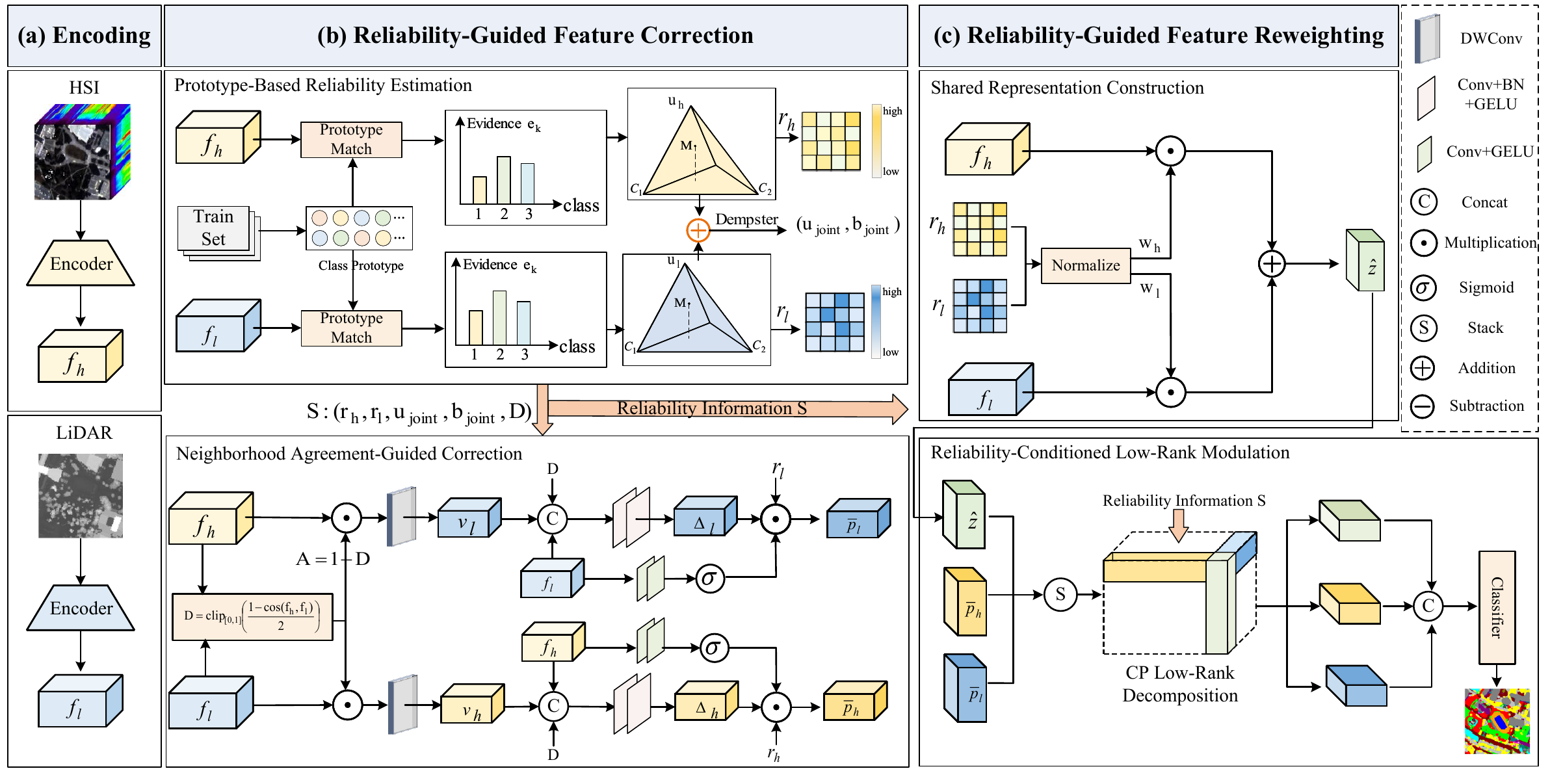}
    \caption{Overall architecture of ReDF.
(a) Modality encoders extract features from aligned
HSI and LiDAR patches.
(b) Class-prototype matching provides modality reliability
estimates, while neighborhood agreement guides bidirectional
cross-modal feature correction.
(c) The modality features and reliability estimates are used
to construct a reliability-weighted shared representation.
The shared and modality-specific residual descriptors,
together with the reliability information from (b), drive
CP low-rank modulation to reweight the residual descriptors.
The shared descriptor is preserved and concatenated with
the reweighted residual descriptors for final classification.}
    \label{fig:framework}
\end{figure*}

As illustrated in Fig.~\ref{fig:framework}, ReDF uses spatially aligned
HSI and LiDAR patches to predict the land-cover class of the central pixel. Let $K$ be the number of classes, and let
$H\times W$ denote the patch size. Two modality-specific shallow feature encoders map the input
patches to feature maps $\mathbf{F}_h,\mathbf{F}_l\in
\mathbb{R}^{d\times H\times W}$, where $d$ denotes the number
of feature channels. Each encoder consists of a $3\times3$
convolution followed by batch normalization and ReLU.
The framework connects spatial feature correction with descriptor-level
reweighting through a shared reference and two modality-specific residuals.
RGFC estimates local reliability from class-prototype evidence, constructs
a reliability-weighted shared reference, and learns neighborhood-supported
correction residuals around this reference. RGFR pools the shared reference
and these same residuals, then adjusts the residual contributions using both
descriptor content and reliability-related statistics. The shared descriptor
passes directly to the classifier. Thus, reliability first controls the
local residual amplitudes and subsequently conditions their channel-wise
reweighting for classification.

\subsection{Reliability-Guided Feature Correction}
\label{subsec:feature_correction}

RGFC uses two complementary cues to construct corrected feature
representations. Prototype evidence measures class-related support within
each modality, whereas cross-modal agreement weights source features
before neighborhood aggregation. The estimated reliability determines
the shared reference and scales each modality's correction residual.

\subsubsection{Prototype-Based Evidence and Reliability}

The encoded feature maps are further transformed by modality-specific
projection layers, yielding
$\mathbf{f}_h,\mathbf{f}_l\in\mathbb{R}^{d\times H\times W}$.
Each projection consists of a $1\times1$ convolution followed by batch
normalization and GELU, preserving the channel count and spatial dimensions.

For each modality $m\in\{h,l\}$ and land-cover class $k$, the prototype
bank stores up to $M_m$ normalized feature vectors.
The $j$th prototype is denoted by
$\mathbf{P}_m^{k,j}\in\mathbb{R}^{d}$, where $j$ indexes the
prototype slots.
The set $\mathcal{J}_m^k$ contains the initialized slots for class $k$
in modality $m$; uninitialized slots are excluded from computation.

During training, the prototype bank is updated without backpropagation
after each optimizer step, using recomputed projected features.
For sample $i$ in a batch, the projected feature map is spatially
pooled and normalized:
\begin{equation}
\mathbf{g}_{m,i}
=\mathcal{N}\!\left(
\operatorname{GAP}(\mathbf{f}_{m,i})
\right).
\label{eq:prototype_sample}
\end{equation}
Here, $\operatorname{GAP}$ denotes global average pooling, and
$\mathcal{N}(\mathbf{x})
=\mathbf{x}/\max(\|\mathbf{x}\|_2,\epsilon)$ denotes $L_2$
normalization, with a small positive constant $\epsilon$ for
numerical stability.
The resulting $\mathbf{g}_{m,i}$ is a $d$-dimensional descriptor
assigned to a class using the sample's center-pixel label.

Within each class, descriptors are assigned to the active prototype
with the highest cosine similarity.
Let $\mathcal{S}_m^{k,j}$ contain the indices of samples assigned to
slot $j$ of class $k$ in modality $m$.
For a nonempty assignment set, the prototype is updated as
\begin{equation}\label{eq:proto_update}
\begin{aligned}
\bar{\mathbf g}_m^{k,j} &= \mathcal N\!\Big(\tfrac{1}{|\mathcal S_m^{k,j}|}\textstyle\sum_{i\in\mathcal S_m^{k,j}}\mathbf g_{m,i}\Big),\\
\mathbf P_m^{k,j} &\leftarrow \mathcal N\!\big((1-\mu)\mathbf P_m^{k,j}+\mu\,\bar{\mathbf g}_m^{k,j}\big).
\end{aligned}
\end{equation}
The update rate $\mu$ controls adaptation to the current batch.
Initialization and stale-slot refresh use descriptors from the
corresponding class.
The bank thus provides class references learned from patch descriptors
with center-pixel labels.

A class contributes to reliability estimation only when both modalities
have at least one active prototype for that class.
We denote the set of such classes by $\mathcal{V}$ and its size by
$K_v=|\mathcal{V}|$.
Evidence is computed separately at each spatial position; spatial indices
are omitted for brevity.
At each position, the projected feature is normalized across channels as
$\boldsymbol{\phi}_m=\mathcal{N}(\mathbf{f}_m)$.
Its evidence for class $k\in\mathcal{V}$ is
\begin{equation}
e_m^k
=\operatorname{softplus}\!\left(
\frac{
\displaystyle\max_{j\in\mathcal{J}_m^k}
\langle\boldsymbol{\phi}_m,\mathbf{P}_m^{k,j}\rangle
}{
\tau_{\mathrm{proto}}
}
\right).
\label{eq:prototype_evidence}
\end{equation}
The inner product measures cosine similarity between the normalized
local feature and each active class prototype.
The best-matching prototype supplies the class score, which is
converted into nonnegative evidence by
$\operatorname{softplus}(x)=\log(1+\exp(x))$.
The prototypes summarize pooled patch features, whereas evidence is
evaluated at individual spatial positions.
The learnable temperature $\tau_{\mathrm{proto}}$ is shared by both
modalities, initialized to $0.1$, and clipped to $[0.01,1]$.

We parameterize a Dirichlet distribution over the valid classes using
$\alpha_m^k=1+e_m^k$.
The additive constant corresponds to a uniform Dirichlet prior with
unit concentration for each valid class.
The total concentration $S_m$, belief mass $b_m^k$, vacuity $u_m$,
and reliability score $r_m$ are
\begin{equation}
\begin{aligned}
S_m &= K_v+\sum_{k\in\mathcal{V}}e_m^k,
&\qquad b_m^k &= \frac{e_m^k}{S_m},\\
u_m &= \frac{K_v}{S_m},
&\qquad r_m &= 1-u_m.
\end{aligned}
\label{eq:belief_uncertainty_reliability}
\end{equation}
These quantities satisfy
$\sum_{k\in\mathcal{V}}b_m^k+u_m=1$.
Vacuity $u_m$ represents uncertainty associated with insufficient evidence.
We use its complement $r_m$ as an evidence-strength-based reliability score.
This score depends on total evidence, rather than calibrated prediction
correctness or how evidence is distributed among competing classes.
For subsequent computations, $r_m$ is clipped to
$[10^{-4},1-10^{-4}]$.
When $K_v=0$, both modality reliabilities are set to $0.5$,
and joint evidence is unavailable.

For $K_v>0$, we also combine the modality-specific belief masses and
vacuities using Dempster's rule.
The conflict mass measures support assigned to different classes by
the two modalities at the same spatial position:
\begin{equation}
\kappa
=
\sum_{\substack{k,k'\in\mathcal{V}\\k\neq k'}}
b_h^k b_l^{k'}.
\label{eq:conflict_mass}
\end{equation}
Here, $b_h^k$ and $b_l^{k'}$ denote the HSI and LiDAR belief masses
for classes $k$ and $k'$, respectively.
The resulting joint belief mass for class $k$ and joint vacuity are
\begin{equation}
\begin{aligned}
b_{\mathrm{joint}}^k
&=
\frac{
b_h^k b_l^k+b_h^k u_l+u_h b_l^k
}{
1-\kappa
},\\
u_{\mathrm{joint}}
&=
\frac{u_h u_l}{1-\kappa}.
\end{aligned}
\label{eq:joint_evidence}
\end{equation}
The numerator of $b_{\mathrm{joint}}^k$ combines matching class support
and terms pairing one modality's class support with the other's vacuity.
The denominator normalizes the nonconflicting mass and is bounded
below by $10^{-6}$ in computation.
Joint vacuity contributes to the reliability statistics used by RGFR.
Joint belief masses determine the class weights in the
prototype-guided feature constraint used during training.

\subsubsection{Shared Reference and Neighborhood-Guided Correction}
\begin{figure*}[!t]
    \centering
    \includegraphics[width=\textwidth,pagebox=cropbox,clip]{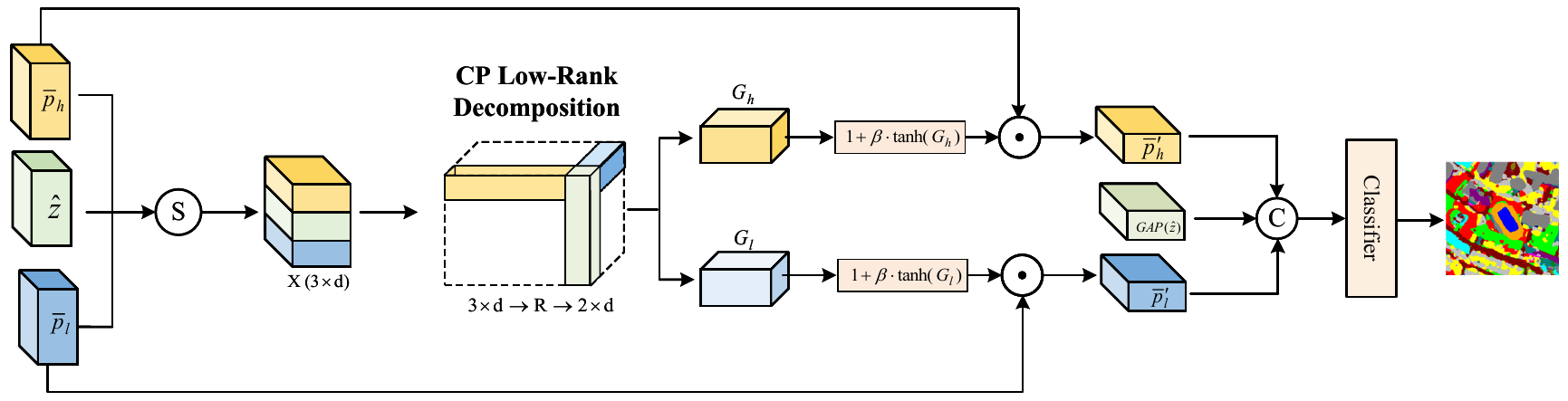}
   \caption{Reliability-Guided Feature Reweighting (RGFR).
Conditioned on reliability information from the preceding
RGFC stage, CP low-rank modulation reweights the
modality-specific residual descriptors, with a learned
scalar $\beta$ controlling the modulation amplitude.
The shared descriptor bypasses reweighting and is
concatenated with the reweighted residuals for classification.
S denotes stacking, C denotes concatenation, and filled-dot
nodes denote element-wise multiplication.}
    \label{fig:rgfr}
\end{figure*}

We first construct a shared reference by weighting the projected
features according to their local reliability:
\begin{equation}
\widehat{\mathbf{z}}
=
\frac{
r_h\mathbf{f}_h+r_l\mathbf{f}_l
}{
r_h+r_l
}.
\label{eq:shared_feature}
\end{equation}
The reliability maps $r_h,r_l\in\mathbb{R}^{1\times H\times W}$
are broadcast across feature channels, and division is performed
element-wise.
This reference favors the modality with stronger local prototype evidence.
Each corrected representation retains a modality-specific residual
around $\widehat{\mathbf{z}}$, allowing it to contribute information
beyond the shared reference.

Strong class evidence alone does not establish local agreement between
the two modalities.
We therefore compute pointwise agreement between their spatially
centered features and use it to weight source features before
neighborhood aggregation.

For each channel, we subtract its spatial mean from $\mathbf{f}_h$
and $\mathbf{f}_l$, yielding the centered feature maps
$\widetilde{\mathbf{f}}_h$ and $\widetilde{\mathbf{f}}_l$.
At each spatial position, cosine similarity is computed across channels
to obtain the disagreement score $D$ and agreement score $A$:
\begin{equation}
D
=\operatorname{clip}_{[0,1]}\!\left(
\frac{
1-\cos(\widetilde{\mathbf{f}}_h,\widetilde{\mathbf{f}}_l)
}{2}
\right),
\qquad
A=1-D.
\label{eq:local_agreement}
\end{equation}
The operator $\operatorname{clip}_{[0,1]}$ restricts its argument to
$[0,1]$, and both $D$ and $A$ have shape $1\times H\times W$.

For target modality $m$ and source modality
$n\in\{h,l\}\setminus\{m\}$, the agreement-weighted neighborhood
aggregate is
\begin{equation}
\mathbf{v}_m
=
\operatorname{DWConv}^{(m)}_{9\times9}
\left(A\odot\mathbf{f}_n\right).
\label{eq:corrector_agg}
\end{equation}
Here, $\odot$ denotes element-wise multiplication, with $A$ broadcast
across feature channels.
The operator $\operatorname{DWConv}^{(m)}_{9\times9}$ denotes a
learnable depthwise convolution for the correction direction targeting
modality $m$.
The two directions use separate kernels, initialized to uniform
averaging, with padding that preserves spatial dimensions.

The residual correction is predicted from the target feature,
the neighborhood aggregate, their difference, and the disagreement map:
\begin{equation}
\boldsymbol{\Delta}_m
=
\mathcal{R}_m\!\left(
[\mathbf{f}_m,\mathbf{v}_m,\mathbf{v}_m-\mathbf{f}_m,D]
\right).
\label{eq:corrector_residual}
\end{equation}
Here, $[\,\cdot\,]$ denotes channel concatenation.
The residual-prediction network $\mathcal{R}_m$ contains a
$1\times1$ convolution followed by a $3\times3$ convolution,
each followed by batch normalization and GELU.

A spatially varying channel-wise amplitude map is separately
predicted from the target feature and disagreement:
\begin{equation}
\mathbf{a}_m
=
\sigma\!\left(
\mathcal{G}_m\!\left(
[\operatorname{sg}(\mathbf{f}_m),D]
\right)
\right).
\label{eq:corrector_gate}
\end{equation}
The network $\mathcal{G}_m$ consists of two $1\times1$ convolutions
with GELU between them.
The sigmoid function $\sigma$ maps its output to
$\mathbf{a}_m\in(0,1)^{d\times H\times W}$.
The stop-gradient operator $\operatorname{sg}$ blocks gradient
propagation through its argument.
Thus, gradients from $\mathcal{G}_m$ do not propagate directly
through its target-feature input.
They can still propagate through $D$, and the target-feature input
to $\mathcal{R}_m$ remains differentiable.

The corrected feature is obtained by adding the reliability-scaled,
channel-gated residual to the shared reference:
\begin{equation}
\widehat{\mathbf{f}}_m
=
\widehat{\mathbf{z}}
+r_m\mathbf{a}_m\odot\boldsymbol{\Delta}_m.
\label{eq:feature_correction}
\end{equation}
Here, $r_m$ is broadcast across channels when scaling the residual.
The two correction directions use separate parameters and the same
shared reference.
For fixed $\mathbf{a}_m$ and $\boldsymbol{\Delta}_m$, decreasing
$r_m$ reduces the residual magnitude relative to
$\widehat{\mathbf{z}}$.
Reliability thus regulates the amount of modality-dependent information
retained around the shared reference.

The residual
$\widehat{\mathbf{f}}_m-\widehat{\mathbf{z}}
=r_m\mathbf{a}_m\odot\boldsymbol{\Delta}_m$
is subsequently pooled and reweighted by RGFR.
This connects local feature correction with the subsequent adjustment
of residual contributions during fusion.
\subsection{Reliability-Guided Feature Reweighting}
\label{subsec:reliability_fusion}

RGFC regulates correction residuals at individual spatial positions.
RGFR subsequently adjusts their pooled contributions to classification
using descriptor content and sample-level reliability statistics
(Fig.~\ref{fig:rgfr}).
It preserves the shared descriptor and applies bounded channel-wise
modulation to the two modality-specific residual descriptors.

\subsubsection{Shared and Modality-Specific Descriptors}

We apply global average pooling to the shared reference and each
corrected feature's residual relative to that reference:
\begin{equation}
\overline{\mathbf{z}}
=\operatorname{GAP}(\widehat{\mathbf{z}}),
\qquad
\overline{\mathbf{p}}_m
=\operatorname{GAP}
\left(\widehat{\mathbf{f}}_m-\widehat{\mathbf{z}}\right).
\label{eq:pooled_descriptors}
\end{equation}
Both $\overline{\mathbf{z}}$ and $\overline{\mathbf{p}}_m$ are
$d$-dimensional column vectors.
The shared descriptor $\overline{\mathbf{z}}$ summarizes the common
reference, whereas
$\overline{\mathbf{p}}_m
=\operatorname{GAP}
(r_m\mathbf{a}_m\odot\boldsymbol{\Delta}_m)$
summarizes the correction residual from RGFC.
Here, modality-specific refers to the residual associated with each
correction direction; no orthogonality or statistical independence
is imposed.
The three descriptors are stacked as rows of
$\mathbf{X}\in\mathbb{R}^{3\times d}$:
\begin{equation}
\mathbf{X}
=
\begin{bmatrix}
\overline{\mathbf{z}}^{\mathsf T}\\
\overline{\mathbf{p}}_h^{\mathsf T}\\
\overline{\mathbf{p}}_l^{\mathsf T}
\end{bmatrix}.
\label{eq:fusion_descriptors}
\end{equation}

\subsubsection{Reliability-Conditioned Low-Rank Modulation}

To condition the modulation on sample-level reliability information,
we aggregate the spatial statistics computed in RGFC.
For a single-channel spatial map $Q$, let $\overline{Q}$ denote its
mean over all $H\times W$ positions.
The conditioning input is
\begin{equation}
\mathbf{s}
=
\left[
\overline{D},
\overline{u}_{\mathrm{joint}},
\overline{r}_h,
\overline{r}_l,
\overline{|r_h-r_l|}
\right]^{\mathsf T}
\in\mathbb{R}^{5}.
\label{eq:reliability_stats}
\end{equation}
Its entries summarize cross-modal disagreement, joint vacuity,
the two modality reliabilities, and their mean absolute difference.
The last entry averages the pointwise absolute reliability difference
between the two modalities.
When joint evidence is unavailable, the joint-vacuity entry is set
to zero as a computational fallback.

A multilayer perceptron (MLP) maps $\mathbf{s}$ to normalized weights
over $K_r$ learned latent states:
\begin{equation}
\boldsymbol{\pi}
=
\operatorname{softmax}
\left(\operatorname{MLP}(\mathbf{s})\right)
\in\mathbb{R}^{K_r}.
\label{eq:state_mixture}
\end{equation}
These states provide learned conditioning components without predefined
semantic labels.
The MLP has widths $5\rightarrow16\rightarrow K_r$, with ReLU
between its two linear layers.
Its final layer is initialized to zero, yielding uniform initial
weights over the latent states.

Let $R$ denote the factorization rank and
$\mathbf{U}\in\mathbb{R}^{K_r\times R}$ a learned matrix with
$L_2$-normalized columns.
The state weights are mapped to an $R$-dimensional conditioning vector:
\begin{equation}
\mathbf{c}_{\mathrm{rank}}
=
\sqrt{R}\,
\mathcal{N}
\left(\mathbf{U}^{\mathsf T}\boldsymbol{\pi}\right).
\label{eq:reliability_condition}
\end{equation}

We parameterize the mapping from descriptor content to gate logits
using a canonical polyadic (CP) factorization with $R$ components.
The conditioning vector $\mathbf{c}_{\mathrm{rank}}$ modulates these
components for each input sample.
Four learned factor matrices implement content projection and gate
reconstruction. The input factors
$\mathbf{A}_{\mathrm{in}}\in\mathbb{R}^{3\times R}$ and
$\mathbf{B}_{\mathrm{in}}\in\mathbb{R}^{d\times R}$
project the three descriptors onto $R$ components.
The output factors
$\mathbf{A}_{\mathrm{out}}\in\mathbb{R}^{2\times R}$ and
$\mathbf{B}_{\mathrm{out}}\in\mathbb{R}^{d\times R}$
reconstruct the two channel-wise gate vectors.
Each factor column is normalized.

Let $\ell$ index the $R$ components, $\nu$ the three input rows,
$c$ the $d$ channels, and $o$ the two output rows.
The content score $\rho_\ell$, conditioned score
$\widetilde{\rho}_\ell$, and gate logit $G_{oc}$ are
\begin{equation}
\begin{aligned}
\rho_\ell
&=
\frac{1}{\sqrt{d}}
\sum_{\nu=1}^{3}\sum_{c=1}^{d}
X_{\nu c}
A_{\mathrm{in},\nu\ell}
B_{\mathrm{in},c\ell},\\
\widetilde{\rho}_\ell
&=
\rho_\ell c_{\mathrm{rank},\ell}w_\ell,\\
G_{oc}
&=
\frac{1}{\sqrt{R}}
\sum_{\ell=1}^{R}
\widetilde{\rho}_\ell
A_{\mathrm{out},o\ell}
B_{\mathrm{out},c\ell}.
\end{aligned}
\label{eq:cp_gating}
\end{equation}
Here, $w_\ell$ is a learned scalar controlling the scale of
component $\ell$.
The first operation aggregates descriptor content, the second applies
reliability conditioning, and the third reconstructs
$\mathbf{G}\in\mathbb{R}^{2\times d}$.
The transposes of its first and second rows are denoted by
$\mathbf{G}_h,\mathbf{G}_l\in\mathbb{R}^{d}$, respectively.
Both the shared and residual descriptors therefore inform the gates,
while only the residual descriptors are reweighted.

A learned scalar $\beta_0$ controls the modulation amplitude through
a fixed upper bound $\beta_{\max}$.
The amplitude and reweighted residual descriptors are
\begin{equation}
\begin{aligned}
\beta
&=\beta_{\max}\sigma(\beta_0),\\
\overline{\mathbf{p}}'_m
&=
\overline{\mathbf{p}}_m\odot
\left(
1+\beta\tanh(\mathbf{G}_m)
\right).
\end{aligned}
\label{eq:private_reweighting}
\end{equation}
The function $\tanh$ is applied element-wise to the gate logits.
The gates depend on the input sample, whereas $\beta$ is shared
across samples.
Each residual channel is multiplied by a factor bounded by
$1-\beta_{\max}$ and $1+\beta_{\max}$.
This modulation adjusts the residual descriptors around an identity
scaling factor of one, while preserving the shared descriptor.

The shared descriptor and the two reweighted residual descriptors
are concatenated and passed to a linear classifier:
\begin{equation}
\begin{aligned}
\mathbf{F}_{\mathrm{fuse}}
&=
[\overline{\mathbf{z}};
\overline{\mathbf{p}}'_h;
\overline{\mathbf{p}}'_l]
\in\mathbb{R}^{3d},\\
\mathbf{z}_{\mathrm{cls}}
&=
\mathbf{W}_{\mathrm{cls}}\mathbf{F}_{\mathrm{fuse}}
+\mathbf{b}_{\mathrm{cls}}.
\end{aligned}
\label{eq:fused_representation}
\end{equation}
Here, semicolons denote vector concatenation,
$\mathbf{W}_{\mathrm{cls}}\in\mathbb{R}^{K\times3d}$ and
$\mathbf{b}_{\mathrm{cls}}\in\mathbb{R}^{K}$ are the classifier
parameters, and $K$ is the number of land-cover classes.
The predicted class is the index of the largest entry in
$\mathbf{z}_{\mathrm{cls}}$.
\subsection{Loss Functions}
\label{sec:loss}

The training objective supervises the final prediction and constrains
the intermediate quantities used in reliability estimation, feature
correction, and feature reweighting:
\begin{equation}
\begin{aligned}
\mathcal{L}_{\mathrm{opt}}={}&\mathcal{L}_{\mathrm{cls}}
+\lambda_{\mathrm{cons}}\mathcal{L}_{\mathrm{cons}}
+\lambda_{\mathrm{pf}}\mathcal{L}_{\mathrm{pf}}\\
&+\lambda_{\mathrm{con}}\mathcal{L}_{\mathrm{con}}
+\lambda_{\mathrm{relreg}}\mathcal{L}_{\mathrm{relreg}}
+\lambda_{\mathrm{evreg}}\mathcal{L}_{\mathrm{evreg}}.
\end{aligned}
\label{eq:grouped_training_objective}
\end{equation}
For a batch of $N_{\mathrm{bs}}$ samples, $y_i$ denotes the center-pixel
label and $p$ indexes the $H\times W$ spatial positions of a feature map.
Here, $\mathbb{E}_{i,p}$ denotes the mean over samples and positions,
$d$ is the feature-channel dimension, $\operatorname{sg}$ stops gradients,
and $\epsilon=10^{-8}$. The valid class set $\mathcal{V}$ and active
prototype slots $\mathcal{J}_m^k$ are defined in
Section~\ref{subsec:feature_correction}.

\subsubsection{Classification supervision}
The classification loss acts on the logits $\mathbf{z}_{\mathrm{cls},i}$
computed from the RGFR output $\mathbf{F}_{\mathrm{fuse},i}$ in
\eqref{eq:fused_representation}:
\begin{equation}
\mathcal{L}_{\mathrm{cls}}
=-\frac{1}{N_{\mathrm{bs}}}\sum_{i=1}^{N_{\mathrm{bs}}}
\log\left[\operatorname{softmax}
\left(\mathbf{z}_{\mathrm{cls},i}\right)\right]_{y_i}.
\label{eq:classification_loss}
\end{equation}
It provides center-pixel supervision to the complete prediction path,
including both RGFC and RGFR.

\subsubsection{Corrected-feature consistency}
This term acts on the corrected feature maps
$\widehat{\mathbf{f}}_h$ and $\widehat{\mathbf{f}}_l$ produced by RGFC
in \eqref{eq:feature_correction}. Define
$w_{i,p}=\operatorname{sg}(\min(r_{h,i,p},r_{l,i,p}))$, using the
reliabilities from \eqref{eq:belief_uncertainty_reliability}. Then
\begin{equation}
\mathcal{L}_{\mathrm{cons}}
=\frac{\displaystyle\sum_{i,p}w_{i,p}
\left[0.5-\cos\left(\widehat{\mathbf{f}}_{h,i,p},
\widehat{\mathbf{f}}_{l,i,p}\right)\right]_+}
{\displaystyle\sum_{i,p}w_{i,p}+\epsilon},
\label{eq:feature_consistency_loss}
\end{equation}
where $[x]_+=\max(x,0)$ and cosine similarity is evaluated over channels.
The loss encourages agreement where both modalities have high reliability,
while imposing no further consistency penalty once the similarity reaches
$0.5$. Detaching $w_{i,p}$ prevents this weighting path from reducing the
loss by lowering reliability; gradients can still pass through the
corrected features, which themselves depend on reliability.

\subsubsection{Prototype-guided feature constraint}
This constraint acts on the projected modality features $\mathbf{f}_m$
and the shared reference $\widehat{\mathbf{z}}$ in
\eqref{eq:shared_feature}, before the correction residuals are added.
For $k\in\mathcal{V}$, define the joint class prototype as
\begin{equation}
\begin{aligned}
\overline{\mathbf{P}}_m^k
&=\frac{1}{|\mathcal{J}_m^k|}
\sum_{j\in\mathcal{J}_m^k}\mathbf{P}_m^{k,j},\\
\mathbf{P}_{\mathrm{joint}}^k
&=\mathcal{N}\!\left(
\frac{\overline{\mathbf{P}}_h^k+\overline{\mathbf{P}}_l^k}{2}
\right).
\end{aligned}
\label{eq:joint_reference_prototype}
\end{equation}
The joint beliefs $b_{\mathrm{joint}}^k$ from
\eqref{eq:joint_evidence} provide spatially varying class weights:
\begin{equation}
q_{i,p}^{k}
=\left[\operatorname{softmax}_{k\in\mathcal{V}}\!\left(
\frac{\log\bigl(\max(b_{\mathrm{joint},i,p}^{k},\epsilon)\bigr)}
{\tau_{\mathrm{post}}}\right)\right]_k,
\label{eq:joint_class_weight}
\end{equation}
where $\tau_{\mathrm{post}}$ is learned, initialized to $0.5$, and
clipped to $[0.01,2]$. The constraint is
\begin{equation}
\begin{aligned}
\mathcal{L}_{\mathrm{recon}}
&=\mathbb{E}_{i,p}\!\left[
\sum_{m\in\{h,l\}}\frac{r_{m,i,p}}{d}
\left\|\mathbf{f}_{m,i,p}-\widehat{\mathbf{z}}_{i,p}\right\|_2^2
\right],\\
\mathcal{L}_{\mathrm{align}}
&=\mathbb{E}_{i,p}\!\left[
\sum_{k\in\mathcal{V}}q_{i,p}^{k}
\left\|\widehat{\mathbf{z}}_{i,p}-\mathbf{P}_{\mathrm{joint}}^k
\right\|_2^2\right],\\
\mathcal{L}_{\mathrm{pf}}
&=\mathcal{L}_{\mathrm{recon}}
+\lambda_{\mathrm{align}}\mathcal{L}_{\mathrm{align}}.
\end{aligned}
\label{eq:prototype_feature_constraint}
\end{equation}
The reconstruction term penalizes reliability-weighted deviations from
the shared reference, with $1/d$ averaging over channels. The alignment
term associates that reference with the joint class prototypes according
to the local joint beliefs. It is a weighted squared-distance constraint.
Both the reliability weights and the class weights remain differentiable
in this loss; the prototypes are updated by the no-gradient procedure in
\eqref{eq:proto_update}.

\subsubsection{Discrimination of fused descriptors}
The supervised contrastive loss acts on
$\mathbf{h}_i=\mathcal{N}(\mathbf{F}_{\mathrm{fuse},i})$, the normalized
descriptor after RGFR in \eqref{eq:fused_representation}.
Let $\mathcal{P}(i)=\{j\ne i:y_j=y_i\}$ and
$\mathcal{I}_+=\{i:|\mathcal{P}(i)|>0\}$.
To adjust the contribution of similar-class negatives, the contrastive
branch uses the first prototype slot of each modality:
\begin{equation}
\mathbf{C}^k=\mathcal{N}\!\left(
\mathcal{N}(\mathbf{P}_h^{k,1})+
\mathcal{N}(\mathbf{P}_l^{k,1})\right).
\label{eq:contrastive_class_prototype}
\end{equation}
For $y_i\ne y_j$ with both labels in $\mathcal{V}$, set
$\chi_{ij}=\max(0,\langle\mathbf{C}^{y_i},\mathbf{C}^{y_j}\rangle)$;
otherwise, set $\chi_{ij}=0$. The pair score and per-anchor loss are
\begin{equation}
\begin{aligned}
\widetilde{s}_{ij}
&=\frac{\mathbf{h}_i^{\mathsf T}\mathbf{h}_j+\delta\chi_{ij}}{T},\\
\xi_i
&=-\frac{1}{|\mathcal{P}(i)|}\sum_{j\in\mathcal{P}(i)}
\log\frac{\exp(\widetilde{s}_{ij})}
{\sum_{t\ne i}\exp(\widetilde{s}_{it})},
\end{aligned}
\label{eq:contrastive_per_anchor}
\end{equation}
where $T=0.07$ and $\delta=0.15$. The additional score increases the
denominator contribution of negatives whose class prototypes are similar,
placing greater emphasis on separating these classes.
Each anchor is weighted by
$\omega_i=\operatorname{sg}(\mathbb{E}_p[
\min(r_{h,i,p},r_{l,i,p})])$, giving
\begin{equation}
\mathcal{L}_{\mathrm{con}}
=\frac{|\mathcal{I}_+|}{N_{\mathrm{bs}}}
\frac{\sum_{i\in\mathcal{I}_+}\omega_i\xi_i}
{\sum_{i\in\mathcal{I}_+}\omega_i}.
\label{eq:reliability_contrastive_loss}
\end{equation}
The prefactor accounts for the fraction of anchors with positive samples.
This term is set to zero when $\mathcal{I}_+$ is empty or its total
reliability weight is below $\epsilon$.

\subsubsection{Reliability and evidence regularization}
These two terms act on the reliability-estimation stage of RGFC.
The first combines the reliabilities in
\eqref{eq:belief_uncertainty_reliability} with the cross-modal
disagreement $D$ in \eqref{eq:local_agreement}:
\begin{equation}
\mathcal{L}_{\mathrm{relreg}}
=-\mathbb{E}_{i,p}\!\left[
\operatorname{sg}(D_{i,p})
\sum_{m\in\{h,l\}}r_{m,i,p}\log(r_{m,i,p}+\epsilon)
\right].
\label{eq:reliability_score_loss}
\end{equation}
It applies stronger reliability regularization at positions with greater
disagreement. The penalty favors scores near the ends of the admissible
range over intermediate scores; it does not specify which modality should
be more reliable. Since $r_h+r_l$ is not constrained to equal one, this
term is not an entropy over modality-selection probabilities.
The disagreement weight is detached, while gradients pass through the
reliability scores.

The evidence regularizer acts directly on the modality-specific class
evidence $e_m^k$ in \eqref{eq:prototype_evidence}.
For a sample with $y_i\in\mathcal{V}$, define
\begin{equation}
\widetilde{\alpha}_{m,i,p}^{k}
=1+\bigl(1-\mathbf{1}[k=y_i]\bigr)e_{m,i,p}^{k},
\qquad k\in\mathcal{V}.
\label{eq:masked_evidence_parameters}
\end{equation}
The target-class concentration is fixed to one so that this regularizer
penalizes non-target evidence without directly penalizing target evidence.
With $N_v=|\{i:y_i\in\mathcal{V}\}|$, the loss is
\begin{equation}
\begin{aligned}
\mathcal{L}_{\mathrm{evreg}}
={}&\frac{1}{2N_vHW}
\sum_{m\in\{h,l\}}\sum_{i:y_i\in\mathcal{V}}\sum_{p=1}^{HW}\\[-2pt]
&\operatorname{KL}\!\left(
\operatorname{Dir}(\widetilde{\boldsymbol{\alpha}}_{m,i,p})
\,\middle\|\,\operatorname{Dir}(\mathbf{1})\right),
\end{aligned}
\label{eq:evidence_regularizer}
\end{equation}
where $\operatorname{Dir}(\mathbf{1})$ is the uniform Dirichlet prior over
the $|\mathcal{V}|$ valid classes. The loss averages over both modalities,
valid samples, and spatial positions, and is zero when $N_v=0$.
The center label is broadcast across each feature map for this term.
This is a supervision surrogate: in mixed-class patches, it can also
penalize evidence for neighboring classes that differ from the center.

When no valid prototypes are available, $\mathcal{L}_{\mathrm{pf}}$ and
$\mathcal{L}_{\mathrm{evreg}}$ are set to zero and $\chi_{ij}=0$.
The remaining feature and classification losses use the fallback
reliability defined in Section~\ref{subsec:feature_correction};
$\mathcal{L}_{\mathrm{relreg}}$ then contributes no gradient because the
reliabilities are constant and $D$ is detached.
The default weights are $\lambda_{\mathrm{cons}}=0.5$,
$\lambda_{\mathrm{pf}}=0.3$, $\lambda_{\mathrm{align}}=0.5$,
$\lambda_{\mathrm{con}}=0.1$, $\lambda_{\mathrm{relreg}}=0.3$, and
$\lambda_{\mathrm{evreg}}=0.02$; the sensitivity to $\lambda_{\mathrm{pf}}$ and $\lambda_{\mathrm{evreg}}$ is examined in
Section~\ref{sec:parameter_analysis}.
\section{Experiments}
\label{sec:experiments}

We evaluate ReDF on three public HSI--LiDAR datasets.
We first introduce the datasets and experimental settings, followed
by a comparison with existing classification methods.
We then conduct ablation studies to examine the contributions of
the proposed modules.
Finally, we visualize the classification results and learned
representations, and analyze sensitivity to the main hyperparameters.

\subsection{Datasets}
\label{sec:datasets}

\noindent\emph{1) Houston2013:}
This urban dataset was collected over the University of Houston campus and surrounding areas in 2012 and released for the 2013 IEEE GRSS Data Fusion Contest~\cite{debes2014houston}. It contains a 144-band HSI covering 0.38--1.05~$\mu$m and a co-registered LiDAR digital surface model. Both modalities have a spatial resolution of 2.5~m and cover $349\times1905$ pixels. The ground-truth map contains 15,029 labeled pixels from 15 classes.

\noindent\emph{2) MUUFL:}
The MUUFL dataset was acquired over the Gulf Park campus of the University of Southern Mississippi in November 2010~\cite{gader2013muufl}. After noisy bands are removed, the HSI contains 64 bands covering 0.38--1.05~$\mu$m. The accompanying LiDAR data provide elevation information. The scene contains $325\times220$ pixels, with 53,687 labeled pixels from 11 classes~\cite{du2017muufl_labels}.

\noindent\emph{3) Trento:}
This dataset covers a rural area in southern Trento, Italy~\cite{rasti2017trento}. It comprises a 63-band HSI covering 0.42--0.99~$\mu$m and co-registered LiDAR data at a spatial resolution of 1~m. The scene contains $166\times600$ pixels and 30,214 labeled pixels from six classes.

\begin{figure*}[!t]
    \centering
    \includegraphics[width=1\textwidth]{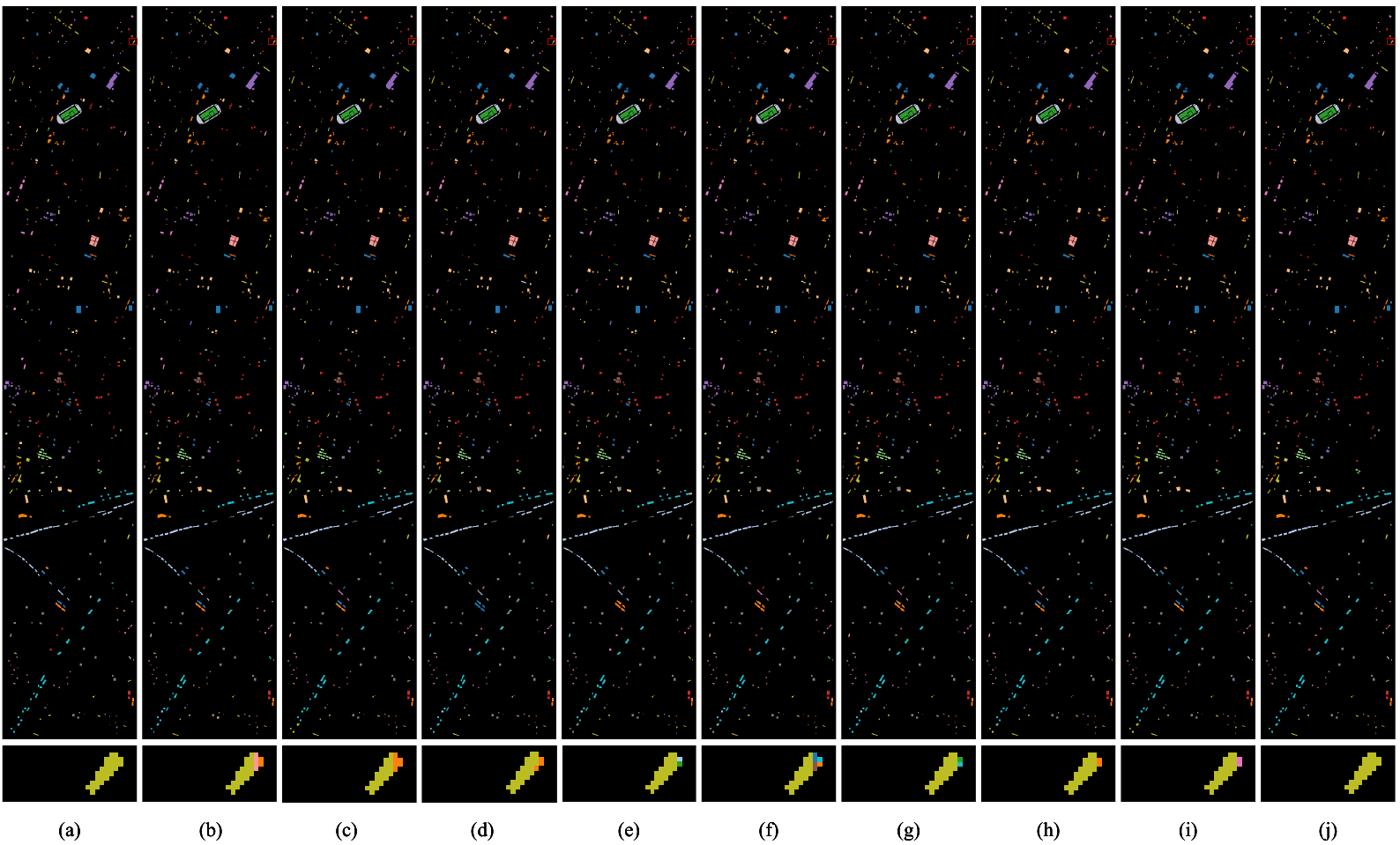}
    \caption{Comparative classification maps for the Houston2013 dataset.
    (a) Ground-truth map.
    (b) CoupledCNN (94.57\%). (c) HCT (93.06\%). (d) DSHFNet (90.23\%). (e) HFSNet (90.35\%).
    (f) DCU-Net (92.93\%). (g) MSAF (92.89\%). (h) ACFFN (95.50\%). (i) mPMCL (94.31\%).
    (j) ReDF (95.82\%). Values in parentheses are the ten-run mean OA from the corresponding table, not the OA of the individual displayed map.}
\label{fig:houston_maps}
\end{figure*}

\begin{figure*}[!t]
    \centering
    \includegraphics[width=1\textwidth]{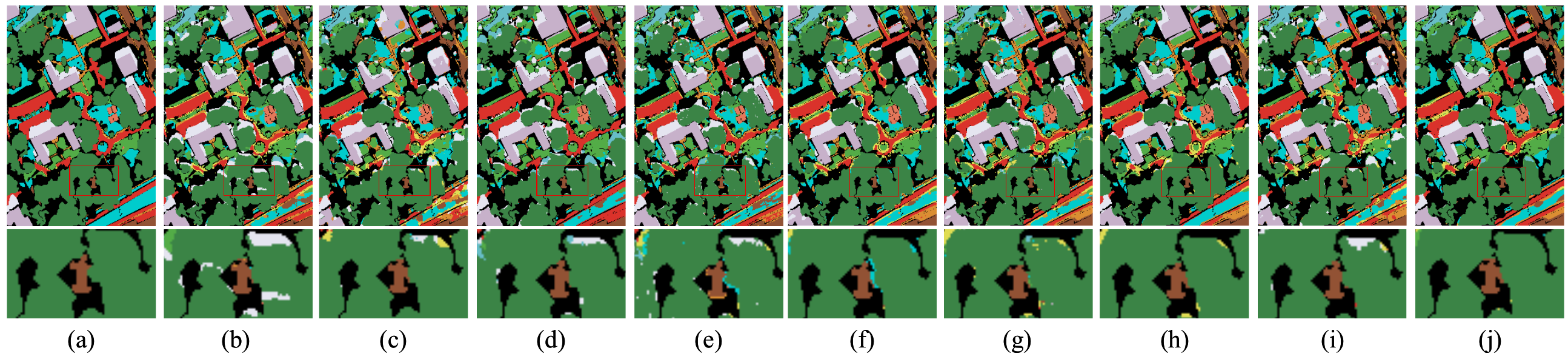}
    \caption{Comparative classification maps for the MUUFL dataset.
    (a) Ground-truth map.
    (b) CoupledCNN (82.53\%). (c) HCT (78.80\%). (d) DSHFNet (79.58\%). (e) HFSNet (82.58\%).
    (f) DCU-Net (83.25\%). (g) MSAF (81.09\%). (h) ACFFN (83.90\%). (i) mPMCL (82.78\%).
    (j) ReDF (84.91\%). Values in parentheses are the ten-run mean OA from the corresponding table, not the OA of the individual displayed map.}
\label{fig:muufl_maps}
\end{figure*}

\begin{figure*}[!t]
\centering

\begin{minipage}{\textwidth}
\centering
\captionof{table}{Mean Classification Accuracies (\%) over Ten Runs on the Trento Dataset}
\label{tab:trento}
\resizebox{\linewidth}{!}{%
\begin{tabular}{clccccccccc}
\toprule
Class No. & Class (Train/Test) & CoupledCNN & HCT & DSHFNet & HFSNet & DCU-Net & MSAF & ACFFN & mPMCL & ReDF \\
\midrule
1 & Apple Trees (20/4014)  & 98.95 & 98.28 & \textbf{99.87} & 97.74 & 98.87 & 99.42 & 99.15 & 99.12 & 99.54 \\
2 & Buildings (20/2883)    & 96.52 & 93.93 & \textbf{99.12} & 98.06 & 96.85 & 97.24 & 97.96 & 96.14 & 98.59 \\
3 & Ground (20/459)        & 98.01 & 98.16 & 97.66 & 96.77 & \textbf{99.06} & 97.60 & 98.47 & 98.45 & 98.38 \\
4 & Woods (20/9103)        & 99.90 & \textbf{100.00} & 99.84 & 99.82 & \textbf{100.00} & 99.96 & 99.86 & \textbf{100.00} & \textbf{100.00} \\
5 & Vineyard (20/10481)    & 99.99 & 97.95 & 99.90 & 99.33 & 99.55 & 99.89 & \textbf{100.00} & 99.15 & 99.99 \\
6 & Roads (20/3154)        & 96.98 & 93.19 & 94.03 & 96.47 & 97.87 & 97.08 & 97.22 & 93.15 & \textbf{97.93} \\
\midrule
  & OA (\%)    & 99.15 & 97.74 & 99.15 & 98.81 & 99.15 & 99.27 & 99.33 & 98.47 & \textbf{99.56} \\
  & AA (\%)    & 98.40 & 96.92 & 98.40 & 98.03 & 98.70 & 98.53 & 98.78 & 97.67 & \textbf{99.07} \\
  & Kappa (\%) & 98.86 & 96.98 & 98.87 & 98.41 & 98.87 & 99.02 & 99.10 & 97.96 & \textbf{99.41} \\
\bottomrule
\end{tabular}
}
\end{minipage}

\par\vspace{8pt}

\includegraphics[width=\textwidth]{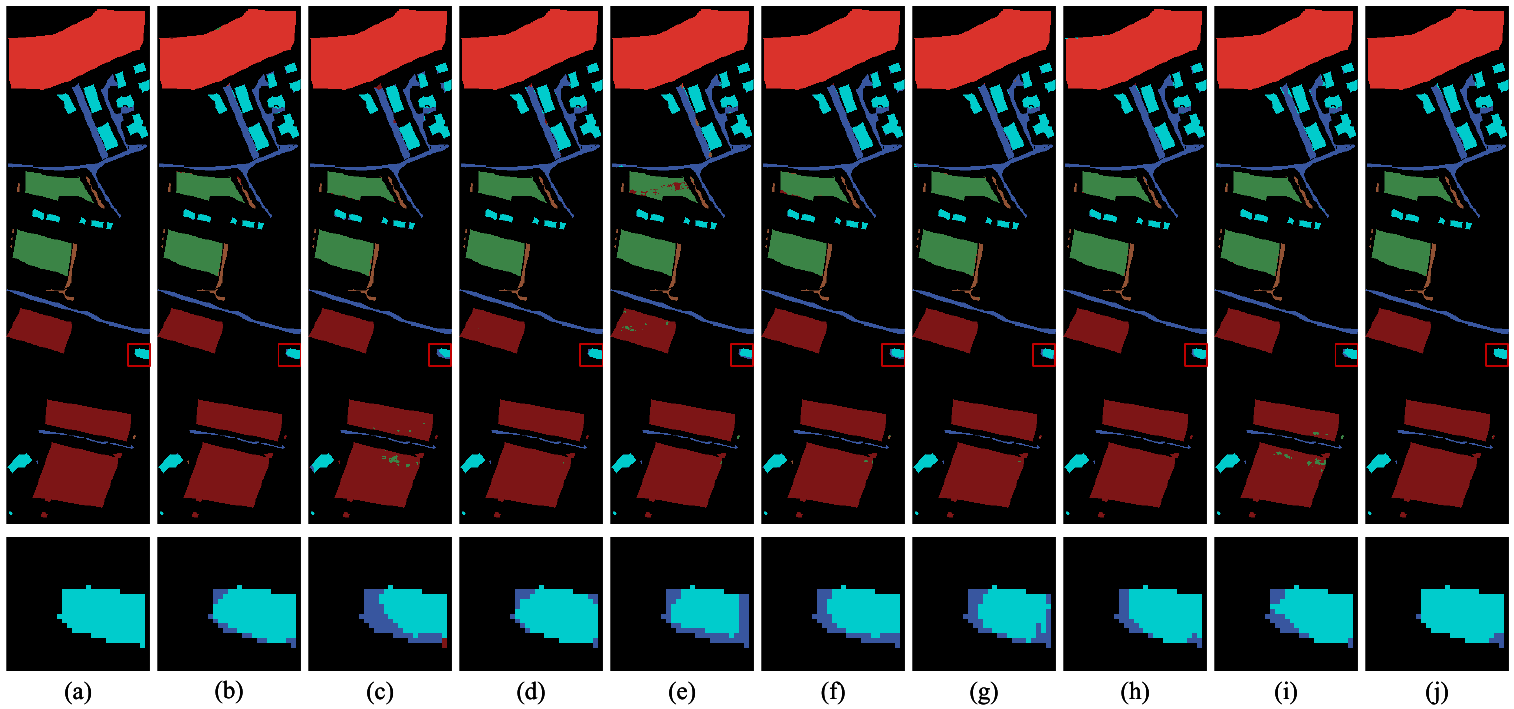}

\caption{Comparative classification maps for the Trento dataset.
(a) Ground-truth map.
(b) CoupledCNN (99.15\%). (c) HCT (97.74\%).
(d) DSHFNet (99.15\%). (e) HFSNet (98.81\%).
(f) DCU-Net (99.15\%). (g) MSAF (99.27\%).
(h) ACFFN (99.33\%). (i) mPMCL (98.47\%).
(j) ReDF (99.56\%). Values in parentheses are the ten-run mean OA
from the corresponding table, not the OA of the individual displayed map.}
\label{fig:trento_maps}

\end{figure*}

For each dataset, we use ten predefined training--test partitions, with 20 training pixels per class and the remaining labeled pixels used for testing. Class-wise sample counts are provided in Tables~\ref{tab:houston}--\ref{tab:trento}, and the ground-truth maps are shown in Figs.~\ref{fig:houston_maps}--\ref{fig:trento_maps}.

\subsection{Experimental Settings}
\label{sec:settings}
\begin{table*}[!t]
\centering

\caption{Mean Classification Accuracies (\%) over Ten Runs on the Houston2013 Dataset}
\label{tab:houston}

\begin{tabular}{clccccccccc}
\toprule
Class No. & Class (Train/Test) & CoupledCNN & HCT & DSHFNet & HFSNet & DCU-Net & MSAF & ACFFN & mPMCL & ReDF \\
\midrule
1  & Healthy Grass (20/1231)   & \textbf{94.35} & 93.89 & 90.60 & 93.82 & 92.92 & 89.83 & 93.41 & 94.16 & 94.24 \\
2  & Stressed Grass (20/1234)  & 95.75 & 96.58 & 95.85 & 93.47 & 95.64 & 95.91 & 96.76 & 97.34 & \textbf{97.94} \\
3  & Synthetic Grass (20/677)  & 99.68 & 99.77 & 99.92 & 99.92 & 99.46 & 99.80 & 99.72 & \textbf{99.94} & 99.67 \\
4  & Tree (20/1224)            & 98.21 & 97.54 & 97.59 & 95.80 & \textbf{98.76} & 96.42 & 96.37 & 97.87 & 97.83 \\
5  & Soil (20/1222)            & 99.28 & \textbf{99.99} & 99.55 & 98.97 & 99.05 & 99.29 & 99.62 & \textbf{99.99} & 99.95 \\
6  & Water (20/305)            & \textbf{99.44} & 97.57 & 95.47 & 92.59 & 94.59 & 96.59 & 97.67 & 97.25 & 97.73 \\
7  & Residential (20/1248)     & 95.87 & 89.21 & 88.11 & 89.22 & 93.32 & 93.83 & \textbf{97.40} & 91.57 & 96.29 \\
8  & Commercial (20/1224)      & 88.69 & 82.39 & 82.69 & 84.50 & 85.19 & 86.94 & 92.34 & 86.74 & \textbf{92.89} \\
9  & Road (20/1232)            & 86.93 & 80.99 & 58.07 & 80.02 & 86.49 & 85.21 & \textbf{91.75} & 83.86 & 90.90 \\
10 & Highway (20/1207)         & 92.37 & 96.93 & 90.62 & 80.93 & 87.30 & 87.97 & 97.26 & \textbf{97.64} & 95.61 \\
11 & Railway (20/1215)         & 95.94 & 94.61 & 90.47 & 87.95 & 91.35 & 92.02 & 95.06 & \textbf{96.14} & 94.58 \\
12 & Park Lot I (20/1213)      & 88.46 & 85.80 & \textbf{92.56} & 87.24 & 87.71 & 89.34 & 87.96 & 88.17 & 91.54 \\
13 & Park Lot II (20/449)      & 97.86 & \textbf{98.46} & 94.74 & 83.51 & 97.48 & 96.99 & 96.44 & 97.08 & 96.57 \\
14 & Tennis Court (20/408)     & 99.97 & \textbf{100.00} & 99.36 & 99.97 & 99.38 & 99.50 & \textbf{100.00} & \textbf{100.00} & \textbf{100.00} \\
15 & Running Track (20/640)    & 99.73 & 99.95 & 99.84 & 99.71 & 99.85 & 99.85 & 99.84 & \textbf{99.97} & 99.85 \\
\midrule
   & OA (\%)    & 94.57 & 93.06 & 90.23 & 90.35 & 92.93 & 92.89 & 95.50 & 94.31 & \textbf{95.82} \\
   & AA (\%)    & 95.51 & 94.25 & 91.70 & 91.18 & 93.90 & 93.97 & 96.11 & 95.18 & \textbf{96.37} \\
   & Kappa (\%) & 94.13 & 92.49 & 89.43 & 89.56 & 92.36 & 92.32 & 95.13 & 93.85 & \textbf{95.48} \\
\bottomrule
\end{tabular}
\end{table*}
\begin{table*}[!t]
\centering

\caption{Mean Classification Accuracies (\%) over Ten Runs on the MUUFL Dataset}
\label{tab:muufl}
\begin{tabular}{clccccccccc}
\toprule
Class No. & Class (Train/Test) & CoupledCNN & HCT & DSHFNet & HFSNet & DCU-Net & MSAF & ACFFN & mPMCL & ReDF \\
\midrule
1  & Trees (20/23226)               & 88.33 & 84.38 & 78.48 & 86.08 & \textbf{90.19} & 87.73 & 89.26 & 86.03 & 90.07 \\
2  & Mostly Grass (20/4250)         & 76.99 & 74.00 & \textbf{80.66} & 79.50 & 75.86 & 70.83 & 75.93 & 77.08 & 77.91 \\
3  & Mixed Ground (20/6862)         & 56.33 & 57.99 & 61.22 & 60.99 & 59.14 & 56.33 & 68.03 & \textbf{71.44} & 67.10 \\
4  & Dirt and Sand (20/1806)        & 87.59 & 81.27 & 90.68 & 87.02 & 81.53 & 79.67 & 87.51 & 90.29 & \textbf{91.25} \\
5  & Road (20/6667)                 & 81.48 & 73.29 & \textbf{89.18} & 84.07 & 80.38 & 80.11 & 79.02 & 79.60 & 84.50 \\
6  & Water (20/446)                 & 99.10 & 98.87 & 99.77 & 99.14 & 99.32 & 99.30 & 99.62 & 99.73 & \textbf{99.97} \\
7  & Building Shadow (20/2213)      & 82.57 & 84.32 & \textbf{94.37} & 89.66 & 87.89 & 85.94 & 86.74 & 83.85 & 86.24 \\
8  & Building (20/6220)             & 95.15 & 90.64 & 89.76 & 93.97 & 93.23 & 92.71 & 94.32 & 92.07 & \textbf{95.96} \\
9  & Sidewalk (20/1365)             & 65.71 & 54.93 & 50.13 & 57.24 & \textbf{65.91} & 63.04 & 57.74 & 55.79 & 45.57 \\
10 & Yellow Curb (20/163)           & 80.61 & 74.53 & 41.84 & 80.06 & 77.30 & 71.41 & 77.12 & \textbf{81.04} & 60.98 \\
11 & Cloth Panels (20/249)          & 95.50 & 95.50 & 97.14 & 93.65 & 95.10 & 94.29 & 95.78 & \textbf{97.30} & 94.57 \\
\midrule
   & OA (\%)    & 82.53 & 78.80 & 79.58 & 82.58 & 83.25 & 81.09 & 83.90 & 82.78 & \textbf{84.91} \\
   & AA (\%)    & 82.67 & 79.07 & 79.38 & 82.85 & 82.35 & 80.12 & 82.82 & \textbf{83.11} & 81.28 \\
       & Kappa (\%) & 77.52 & 72.96 & 74.39 & 77.68 & 78.39 & 75.69 & 79.23 & 77.93 & \textbf{80.48} \\
\bottomrule
\end{tabular}
\end{table*}

ReDF is implemented in PyTorch and trained on a workstation
equipped with an Intel Core i9-14900K CPU and an NVIDIA
RTX 4090 GPU. We use the Adam optimizer with weight decay
$10^{-4}$ and a batch size of 64.
The gradient norm is clipped at 1.0.
The learning rate follows a cosine schedule, decreasing to
1\% of its initial value.
Dataset-specific hyperparameters are as follows: the initial learning rate is set to 
$3\times10^{-3}$, $3\times10^{-4}$, and $5\times10^{-3}$ for Houston2013, MUUFL, and Trento, 
respectively, with training lasting 500, 300, and 300 epochs accordingly. 
The batch size is fixed at 64 for all datasets.


We compare ReDF with CoupledCNN~\cite{ref_coupledcnn}, HCT~\cite{ref_hct}, DSHFNet~\cite{ref_dshfnet}, HFSNet~\cite{ref_hfsnet}, DCU-Net~\cite{ref_dcunet}, MSAF~\cite{ref_msaf}, ACFFN~\cite{ref_acffn}, and mPMCL~\cite{ref_mpmcl}. All compared methods use the same data splits and preprocessing as ReDF, with their official implementations adopted.

We report overall accuracy (OA), average accuracy (AA), class-wise accuracy (CA), and the Kappa coefficient. 
Metrics are computed separately for each run and then averaged. All metrics are reported in percentages, and gains are measured in percentage points. 

\subsection{Performance Comparison}
\label{sec:comparison}

Tables~\ref{tab:houston}--\ref{tab:trento} report the mean
classification results over ten runs on the three datasets.
ReDF achieves the highest OA and Kappa on all datasets,
and the highest AA on Houston2013 and Trento.
The following analysis examines the aggregate results,
representative class-wise improvements, and remaining
classification difficulties.

\subsubsection{Houston2013}

Table~\ref{tab:houston} presents the classification results
on Houston2013.
ReDF achieves an OA of 95.82\%, an AA of 96.37\%, and a
Kappa of 95.48\%, ranking first on all three metrics.
Compared with ACFFN, which ranks second on these metrics,
ReDF improves OA, AA, and Kappa by 0.32, 0.26, and
0.35 percentage points, respectively.

Regarding class-wise performance, ReDF achieves the highest
accuracies on Stressed Grass and Commercial, reaching
97.94\% and 92.89\%, respectively.
These exceed the respective best competing results from
mPMCL and ACFFN by 0.60 and 0.55 percentage points.
On Park Lot I, ReDF reaches 91.54\%, improving on ACFFN
by 3.58 percentage points, although DSHFNet achieves the
highest accuracy for this class.
ReDF also shares the highest accuracy of 100.00\%
on Tennis Court.

However, ReDF remains below ACFFN on Residential, Road, and Highway by
1.11, 0.85, and 1.65 percentage points, respectively.
One possible explanation is that patch-level prototype
references do not fully capture local feature variations
within these categories, limiting the benefit of
reliability-guided correction.
Despite these class-specific differences, ReDF achieves
the highest aggregate performance, supporting its
effectiveness for classification in this urban scene.

\subsubsection{MUUFL}

Table~\ref{tab:muufl} reports the classification results
on MUUFL.
ReDF obtains the highest OA and Kappa, reaching 84.91\%
and 80.48\%, respectively.
Compared with ACFFN, the runner-up on both metrics,
the improvements are 1.01 and 1.25 percentage points.
Its AA is 81.28\%, below the highest value of 83.11\%
obtained by mPMCL, indicating that the aggregate gains
are not accompanied by the best average class accuracy.

At the class level, ReDF achieves the highest accuracies
on Dirt and Sand, Water, and Building, at 91.25\%,
99.97\%, and 95.96\%, respectively.
Compared with ACFFN, it improves accuracy on Dirt and Sand,
Road, and Building by 3.74, 5.48, and 1.64 percentage points.
It also reaches 90.07\% on Trees, approaching the highest
accuracy of 90.19\% obtained by DCU-Net.

Nevertheless, Sidewalk and Yellow Curb remain challenging,
with accuracies of 45.57\% and 60.98\%, respectively.
These are 12.17 and 16.14 percentage points below ACFFN.
A possible explanation is that mixed spatial context
within input patches weakens the local cues needed to
distinguish these categories during neighborhood aggregation.
The resulting class-specific errors limit AA, whereas OA
also reflects the unequal numbers of test pixels across
categories.
Despite these limitations, the highest OA and Kappa
demonstrate an overall classification advantage on MUUFL,
while identifying categories that require further improvement.

\subsubsection{Trento}

Table~\ref{tab:trento} presents the classification results
on Trento.
ReDF achieves the highest OA, AA, and Kappa, at 99.56\%,
99.07\%, and 99.41\%, respectively.
Compared with ACFFN, which ranks second on all three metrics,
the improvements are 0.23, 0.29, and 0.31 percentage points.

ReDF achieves the highest accuracy of 97.93\% on Roads,
exceeding ACFFN by 0.71 percentage points and the best
competing result from DCU-Net by 0.06 percentage points.
On Buildings, it reaches 98.59\%, improving on ACFFN by
0.63 percentage points.
ReDF also shares the highest accuracy of 100.00\% on Woods
and reaches 99.99\% on Vineyard, maintaining high accuracy
on these two categories.

Although ReDF leads on the aggregate metrics, DSHFNet
achieves higher accuracies on Apple Trees and Buildings,
while DCU-Net performs best on Ground.
These differences may reflect variation in how effectively
each model captures the class-specific information needed
to distinguish individual land-cover categories.
Several classes are already classified with nearly
100\% accuracy, leaving limited room for further gains.
Within this setting, ReDF achieves a modest improvement
in all three aggregate metrics over the strongest
competing method.

\subsection{Ablation Study}
\label{sec:ablation}

To evaluate the contributions of the proposed components,
we conduct ablation experiments on Houston2013, MUUFL,
and Trento using the same ten training--test partitions.
Table~\ref{tab:ablation_modules} reports the mean OA and
standard deviation for the complete model and its variants.

\subsubsection{Effect of RGFC}

To assess the contribution of reliability-guided feature
correction, we compare the complete model with a variant
without RGFC.
In this variant, the reliability estimator and correctors
are disabled.
The projected features form an arithmetic-mean shared
reference, and the two residual descriptors supplied to
RGFR are set to zero.

As shown in Table~\ref{tab:ablation_modules}, retaining RGFC
improves classification performance on all three datasets.
On Houston2013, OA increases from 95.29\% to 95.82\%.
The largest improvement occurs on MUUFL, where OA rises
from 80.86\% to 84.91\%, corresponding to a gain of
4.05 percentage points.
On Trento, OA increases from 99.37\% to 99.56\%.
These results support the contribution of the complete
feature-correction path.
By constructing a shared reference and retaining
modality-dependent correction residuals, RGFC provides
additional information for subsequent fusion.
The larger gain on MUUFL suggests that this representation
is particularly useful under the evaluated conditions
of that dataset.

\subsubsection{Effect of RGFR}

To examine the contribution of RGFR, we retain RGFC and
replace RGFR with global average pooling of the concatenated
corrected feature maps, followed by a linear classifier.
This replacement directly classifies the pooled corrected
features without the shared--residual descriptor construction
and reliability-conditioned modulation used by RGFR.

The complete model achieves higher mean OA on all three
datasets.
On Houston2013, RGFR increases OA from 95.80\% to 95.82\%,
yielding only a small difference.
On MUUFL, OA increases from 84.22\% to 84.91\%, while
on Trento it increases from 99.49\% to 99.56\%.
The larger improvement on MUUFL is consistent with the
intended role of RGFR in adjusting residual contributions
according to descriptor content and reliability statistics.
Its effect is smaller on Houston2013 and Trento, indicating
that the benefit varies across datasets.
This comparison evaluates the complete RGFR fusion head,
including its descriptor construction, rather than the
low-rank modulation alone.

\begin{table}[t]
\centering
\small
\setlength{\tabcolsep}{3pt}
\caption{Ablation of ReDF (OA\%, mean $\pm$ standard deviation over ten runs). RGFC and RGFR denote feature correction and feature reweighting, respectively. Estimator and Corrector denote the two components within RGFC.}
\label{tab:ablation_modules}
\label{tab:ablation_correction}
\begin{tabular}{lccc}
\toprule
Variant & Houston2013 & MUUFL & Trento \\
\midrule
\textbf{Full model} & $\boldsymbol{95.82 \pm 0.88}$ & $\boldsymbol{84.91 \pm 1.11}$ & $\boldsymbol{99.56 \pm 0.08}$ \\
w/o RGFC      & $95.29 \pm 0.88$ & $80.86 \pm 1.36$ & $99.37 \pm 0.16$ \\
w/o RGFR      & $95.80 \pm 0.77$ & $84.22 \pm 1.00$ & $99.49 \pm 0.12$ \\
w/o Estimator & $95.70 \pm 0.82$ & $84.76 \pm 0.94$ & $99.52 \pm 0.14$ \\
w/o Corrector & $94.77 \pm 0.68$ & $79.17 \pm 1.56$ & $99.32 \pm 0.17$ \\
\bottomrule
\end{tabular}
\end{table}

\subsubsection{Effect of the Reliability Estimator and Corrector}

To further examine the components within RGFC, we separately
disable the reliability estimator and the corrector.
Without the estimator, both modality reliabilities are fixed
at 0.5, the shared reference becomes the arithmetic mean,
and the unavailable joint-vacuity statistic is set to zero.
The correctors and RGFR remain active.
Without the corrector, the reliability estimator is retained,
but both corrected feature maps are replaced by the shared
reference, producing zero residual descriptors for RGFR.

Disabling the estimator reduces OA from 95.82\% to 95.70\%
on Houston2013, from 84.91\% to 84.76\% on MUUFL, and
from 99.56\% to 99.52\% on Trento.
The complete model therefore maintains a higher mean OA
than the fixed-reliability variant on all three datasets.
These results suggest that adapting the shared reference and residual amplitudes
to local evidence strength is beneficial, although the observed gains are small.

Removing the corrector produces larger decreases.
OA falls to 94.77\%, 79.17\%, and 99.32\% on Houston2013,
MUUFL, and Trento, respectively.
The corresponding reductions are 1.05, 5.74, and
0.24 percentage points, with the largest decrease again
observed on MUUFL.
This suggests that the correction path and the
modality-specific residual information it supplies
are important to the final classification.
Neighborhood aggregation allows the predicted residuals
to incorporate complementary source-modality features,
which may help explain the advantage over using the
shared reference alone.
Because this variant also removes the residual inputs
to RGFR, the observed decrease reflects the combined
effect of feature correction and residual availability.

Overall, the complete model achieves the highest mean OA
among the evaluated variants on all three datasets.
The largest changes are associated with removing the
correction path, whereas reliability estimation and
residual reweighting provide smaller, dataset-dependent
improvements.

\begin{figure}[t]
    \centering
    \includegraphics[width=\columnwidth]{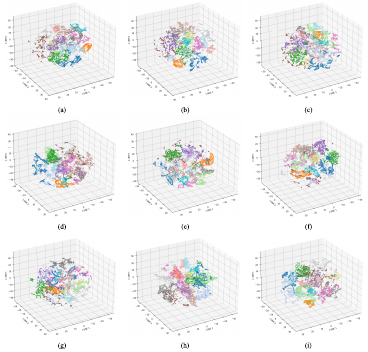}
    \caption{Three-dimensional t-SNE visualization of test features on the
    Houston2013 dataset. (a) CoupledCNN, (b) HCT, (c) DSHFNet,
    (d) HFSNet, (e) DCU-Net, (f) MSAF, (g) ACFFN, (h) mPMCL, and
    (i) ReDF. Colors denote the ground-truth classes.}
\label{fig:tsne_houston}
\end{figure}

\subsection{Parameter Sensitivity Analysis}
\label{sec:parameter_analysis}
\subsubsection{Visual Analysis}
\label{sec:param}
\label{sec:feature_visualization}
 Figure~\ref{fig:tsne_houston} shows three-dimensional t-SNE embeddings of the Houston2013 test features.
For all methods, the central region contains partially overlapping classes, which
reflects the spectral similarity among several urban categories. ReDF forms
several compact and well-isolated clusters, whereas the embeddings of some competing
methods, such as ACFFN, show more interleaving in the central region. Some classes nevertheless remain mixed in the embedding of ReDF, which agrees with
its lower accuracies than ACFFN on Residential, Road, and Highway in Table~I. Since
t-SNE does not preserve global distances, this comparison is qualitative.
\subsubsection{Latent States and Prototype Slots}
\begin{figure}[t]
    \centering
    \includegraphics[width=\columnwidth]{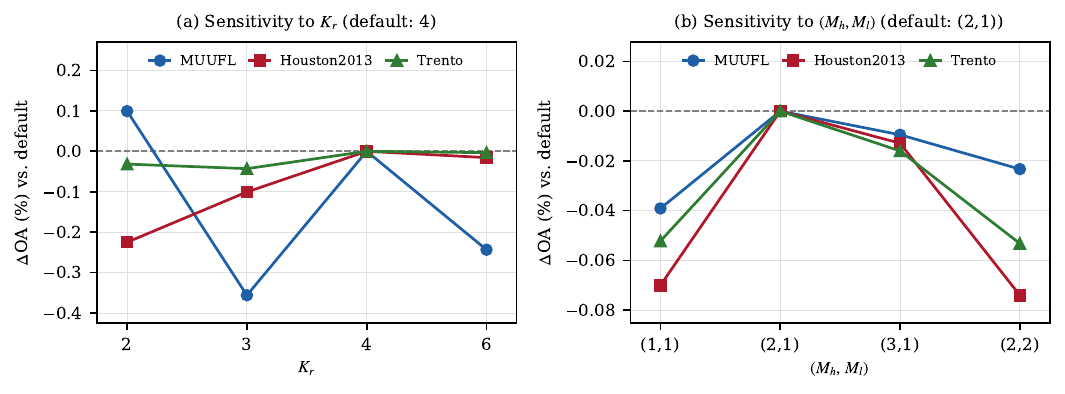}
    \caption{Sensitivity to (a) $K_r$ and (b) $(M_h,M_l)$ on MUUFL,
    Houston2013, and Trento. The vertical axis shows the change in mean OA
    relative to the default setting (gray dashed line, $\Delta$OA $=0$).
    The plotted values are mean differences in percentage points.}
    \label{fig:hparam}
\end{figure}

We examine $K_r\in\{2,3,4,6\}$ and $(M_h,M_l)\in\{(1,1),(2,1),(3,1),(2,2)\}$, varying one configuration at a time while holding the remaining settings fixed. Figure~\ref{fig:hparam} reports the difference in mean OA from the default configuration, so the zero line represents the full model's default setting.

With $K_r=4$, ReDF attains the highest mean OA on Houston2013 and ties for the highest on Trento among the tested values. On MUUFL, $K_r=2$ improves OA by 0.10 percentage points over the default. Across the tested values, the range of mean OA is approximately 0.46 points on MUUFL, 0.22 on Houston2013, and 0.04 on Trento. We use $K_r=4$ as a common configuration across datasets.

The default prototype configuration $(M_h,M_l)=(2,1)$ gives the highest mean OA on all three datasets in the reported sweep. The largest decreases for the alternative configurations are approximately 0.04, 0.07, and 0.05 points on MUUFL, Houston2013, and Trento, respectively. These small differences characterize the tested configurations and do not indicate a large performance advantage from increasing the number of prototype slots.

\subsubsection{Auxiliary Loss Weights}
We vary the prototype-guided feature-constraint weight $\lambda_{\mathrm{pf}}\in\{0.15,0.3,0.6,1.2\}$ on Houston2013 and the evidence-regularization weight $\lambda_{\mathrm{evreg}}\in\{0.005,0.01,0.02,0.04,0.08\}$ on Trento. These experiments examine the weights associated with shared-feature learning and class-evidence regularization, respectively.

\begin{figure}[t]
    \centering
    \includegraphics[width=\columnwidth]{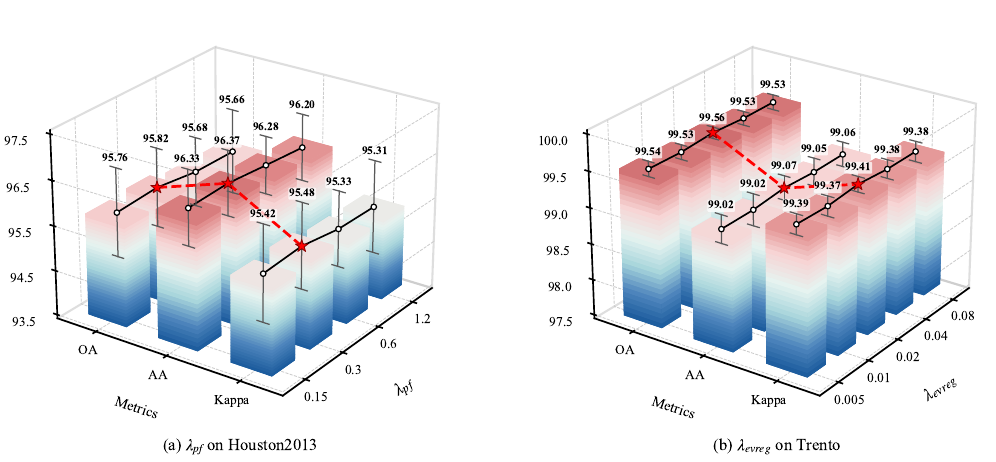}
    \caption{Loss-weight sensitivity: (a) $\lambda_{\mathrm{pf}}$ on Houston2013 and (b) $\lambda_{\mathrm{evreg}}$ on Trento. The vertical axis shows OA, AA, and Kappa (\%); bar color encodes height only. Bars and black markers show the mean over 10 runs; error bars indicate standard deviations. Red stars mark the highest mean for each metric, and the red dashed line connects these points.}
    \label{fig:loss_weight}
\end{figure}

In the reported sweeps, $\lambda_{\mathrm{pf}}=0.3$ and $\lambda_{\mathrm{evreg}}=0.02$ yield the highest mean OA, AA, and Kappa on their respective datasets (Fig.~\ref{fig:loss_weight}). The mean differences across the tested settings are small relative to the displayed run-to-run standard deviations. We retain these values as the default weights; the plots summarize sensitivity within the tested ranges.

\section{Conclusion}
\label{sec:conclusion}

This work presents ReDF, a reliability-driven framework for HSI–LiDAR land-cover classification. ReDF estimates modality reliability from local class-prototype evidence and uses it to guide both cross-modal feature correction and subsequent fusion. RGFC corrects modality features using a reliability-weighted shared reference and neighborhood agreement, while RGFR combines reliability information with descriptor content to reweight modality-specific correction residuals. Under the reported experimental settings, ReDF achieves higher mean OA and Kappa than the compared methods on three public datasets. Ablation results further support the contribution of the complete feature-correction path and show that the benefits of reliability estimation and residual reweighting vary across datasets.
Nevertheless, ReDF remains less effective for categories such as Sidewalk and Yellow Curb on MUUFL. Future work will examine local reliability estimation and feature-correction behavior in patches containing multiple classes and explore finer-grained spatial supervision to improve classification in complex regions.

\bibliographystyle{IEEEtran}
\bibliography{references}

\end{document}